\documentclass[10pt,twocolumn,letterpaper]{article}

\usepackage{multirow}
\usepackage{float}

\usepackage[pagenumbers]{cvpr} 

\usepackage[pagebackref,breaklinks,colorlinks]{hyperref}
\usepackage{tabularx}
\usepackage{pgfplots}
\usepgfplotslibrary{statistics}
\usepackage{pgf,tikz}
\usepackage{mathtools}
\usepackage{amsmath}
\usepackage{bm}

\definecolor{cPLOT0}{RGB}{214,113,176}
\definecolor{cPLOT1}{RGB}{80,150,80}
\definecolor{cPLOT3}{RGB}{165, 124, 27}
\definecolor{cPLOT2}{RGB}{68, 33, 175}
\definecolor{cPLOT5}{RGB}{39, 174, 239}
\definecolor{cPLOT6}{RGB}{179,0,0}

\title{Geometrically Consistent Partial Shape Matching}

\author{Viktoria Ehm$^{1,2}$
\and
Paul Roetzer$^{3}$
\and
Marvin Eisenberger$^{1}$
\and
Maolin Gao$^{1}$
\and
Florian Bernard$^{3}$
\and
Daniel Cremers$^{1,2}$
\\
$^{1}$ TU Munich $^{2}$ Munich Center for Machine Learning $^{3}$ University of Bonn
}

\begin{document}%
\twocolumn[{%
\renewcommand\twocolumn[1][]{#1}%
\maketitle
\begin{center}%
    \centering%
    \captionsetup{type=figure}%
    \hspace*{0.1in}%
   \includegraphics[width=1\textwidth,trim={0 12cm 0 12.7cm},clip]{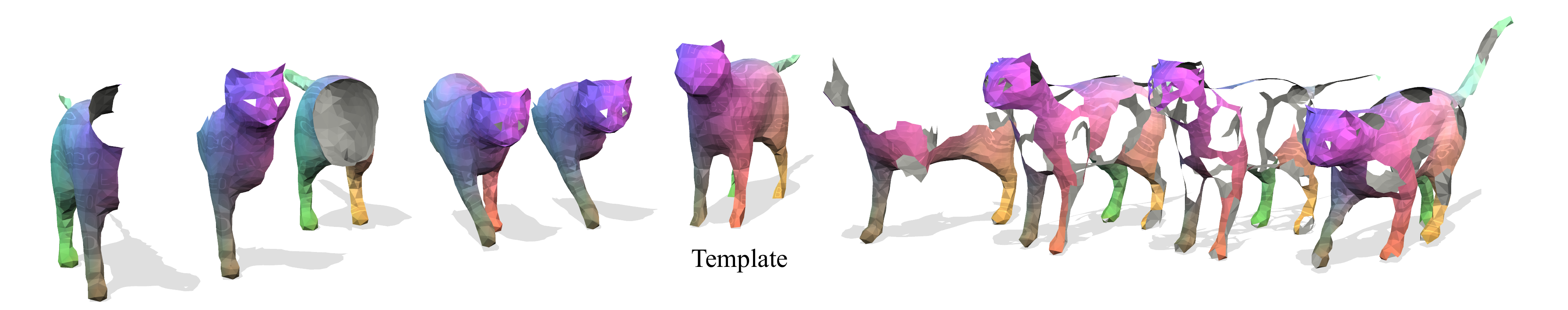}
    \captionof{figure}{We find \textbf{geometrically consistent matchings} between a template shape (middle) and \textbf{partial shapes} in different poses (left and right). The left shows shapes  that are cut by a single clean cut, whereas the right shows shapes with multiple holes in different sizes.}
\end{center}%
}]

\begin{abstract}

Finding correspondences between 3D shapes is a crucial problem in computer vision and graphics, which is for example relevant for tasks like shape interpolation, pose transfer, or texture transfer. An often neglected but essential property of matchings is geometric consistency, which means that neighboring triangles in one shape are consistently matched to neighboring triangles in the other shape. Moreover, while in practice one often has only access to partial observations of a 3D shape (e.g.~due to occlusion, or scanning artifacts), there do not exist any methods that directly address geometrically consistent partial shape matching. In this work we fill this gap by proposing to integrate state-of-the-art deep shape features into a novel integer linear programming partial shape matching formulation. Our optimization yields a globally optimal solution on low resolution shapes, which we then refine using a coarse-to-fine scheme.
We show that our method can find more reliable results on partial shapes in comparison to existing geometrically consistent algorithms (for which one first has to fill missing parts with a dummy geometry).
Moreover, our matchings are substantially smoother than learning-based state-of-the-art shape matching methods.

\end{abstract}  
\section{Introduction}
\label{sec:intro}

Computing correspondences between deformable 3D surfaces is a fundamental challenge in 3D computer vision. 
While there is a large body of literature, to date there is still a significant domain gap that prevents state-of-the-art (SOTA) approaches from being deployed in real world applications. 
Real scans are often subject to several challenges, including fine-scale noise, clutter, occlusion, topological changes, and partial views.

In recent times, a significant percentage of new approaches that address shape correspondence fall in the framework of geometric deep learning.
A major strength of such approaches is that they obtain flexible local feature representations which widely outperform shape descriptors that are manually defined by human experts. 
On the other hand, existing deep learning approaches often yield non-bijective correspondence maps that lack a meaningful geometric interpretation and are prone to considerable local noise.
This issue is detrimental for many applications such as normal estimation,
pose transfer, and texture mapping, where the local fidelity of a predicted map plays a crucial role. 
This problem is, in part, exacerbated by popular benchmarks, which mostly focus on the geodesic error accuracy metric, which is disproportionally impacted by outliers and global mismatches. 
By comparison, local noise is only reflected to a lesser extent.

In this work, we propose a novel solution to the problem of matching partial shapes. 
Our approach combines deep local feature embeddings with classical Integer Linear Programming, to get the best of both worlds.
By ensuring geometric consistency in our Integer Linear Program we can reduce local noise in the solutions, so that matchings can for example be used for shape interpolation between partial and full shapes. 
We solve our approach to global optimality on a coarse level and then gradually increase to higher resolutions.
We summarize our main contributions as follows:

\begin{itemize}
    \item We propose a novel Integer Linear Program optimization approach for partial shape matching. Our formulation leads to geometrically consistent correspondences for this challenging setting, \ie preserves the local triangle neighborhood structure.
    \item We devise an iterative coarse-to-fine scheme that allows to expand globally optimal low-resolution maps to higher resolutions in a pruned search space.
    \item We introduce a manifold version of the SHREC16 partial dataset, which is the standard benchmark for partial shape correspondence in the literature\footnote{We will publish the dataset to the research community upon acceptance.}. 
    \item We show that our method leads to state-of-the-art performance regarding smoothness and accuracy in partial matching, both in terms of cuts (missing parts) and holes.   
\end{itemize}
\section{Related Work}
\label{sec:relatedWork}
A comprehensive review of the field of 3D shape matching is beyond the scope of this paper, so we only focus on the most related works and refer interested readers to recent surveys~\cite{van2011survey,sahilliouglu2020recent}.

\paragraph{Global Shape Matching.}
Matching pairs of shapes is a well-studied problem and shortest path algorithms can be used to efficiently find  global optima in various 2D settings~\cite{sakoe1978dynamic,schmidt2007fast,michel2011scale,lahner2016efficient,roetzer2023conjugate}. Interestingly, just by going one dimension higher, namely 3D shape matching, the problem becomes considerably harder and the solution is a minimal surface (instead of a shortest path) in the four dimensional product manifold. An elegant \emph{integer linear programming} (\textbf{ILP}) formulation has been proposed in~\cite{windheuser2011geometrically, windheuser2011large, Schmidt-et-al-14}, which is guaranteed to produce continuous and orientation-preserving correspondences. However, it requires watertight meshes as input, which makes it necessary to pre-process  \emph{partial} shapes by closing holes with a dummy geometry~\cite{windheuser2011geometrically, Schmidt-et-al-14, roetzer2022scalable}. Moreover due to the exponential size of the combinatorial solution space, the global optimum can not be obtained even for moderately large problems. Recently, a \emph{mixed integer programming} (\textbf{MIP}) formulation has been proposed in~\cite{bernard2020mina}, which, however, can only be optimized to global optimality for a small set of sparse keypoints.

Another popular approach is the functional map pipeline~\cite{ovsjanikov2012functional}, which elegantly transfers the matching problem from the spatial domain (matching vertices) to the functional domain (matching functions). As a result, the optimization becomes a linear least-squares problem, which is known to be solvable to global optimality efficiently. Consequently, it has been extended to various variants, including the partial setting~\cite{rodola2017partial, attaiki2021dpfm}, non-isometric shapes~\cite{eisenberger2020}, or shape collections~\cite{CZO_SGP20, gao2021multi}. Yet, obtaining geometrically consistent matchings from functional map solutions is still an open problem.

\paragraph{Partial Shape Matching}~receives particular attention due to its high practical relevance. In comparison with matching full shapes, partial shapes exhibit severe non-isometries and irregularity, which makes it a hard problem to solve. 
In~\cite{rodola2017partial}, an extension of the functional map framework to partial cases is introduced by perturbation analysis of the underlying \emph{Laplace-Beltrami Operator} (\textbf{LBO}), where the partial functional map together with a (soft) indicator function to localize the overlapping parts are optimized. Follow-up works have successfully extended this idea to clutter~\cite{Cosmo2016}, multiple shapes~\cite{Litany2016}, fully spectral formalisms~\cite{litany2017fully} and the deep learning regime~\cite{attaiki2021dpfm,cao2023unsupervised}. 
One drawback many methods inherit from functional maps is that 
the point-wise correspondence has to be recovered using post-processing techniques, for which it is not straight-forward to guarantee desired structure properties, such as bijectivity and smoothness~\cite{vestner2017pmf, ren2018continuous}.
In contrast, our method can incorporate such desirable properties via constraints in the optimization
and operates directly in the spatial domain. With that,  point-wise correspondences can be obtained without any post-processing steps.

\begin{figure*}
    \centering
    \includegraphics[width=1\linewidth]{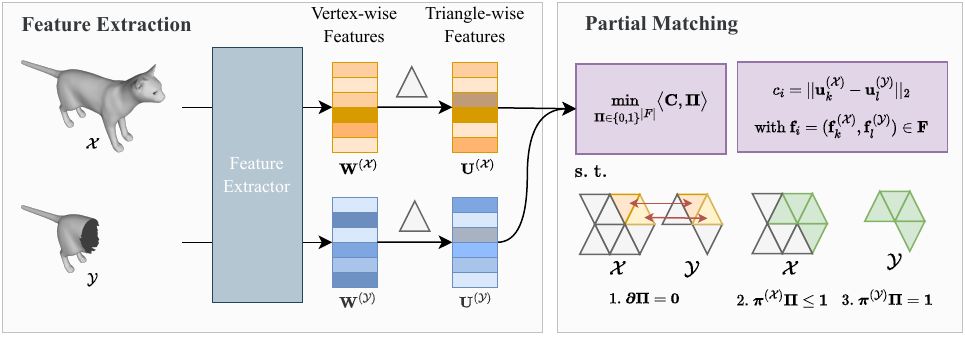}
    \caption{We show an \textbf{overview} of our   \textbf{geometrically consistent partial shape matching} approach. After extracting vertex-wise features $\mathbf{W}^{(\mathcal{X})}, \mathbf{W}^{(\mathcal{Y})}$ for full shape $\mathcal{X}$ and partial shape $\mathcal{Y}$, we convert them to triangle-wise features $\mathbf{u}_k^{(\mathcal{X})} \in \mathbf{U}^{(\mathcal{X})}$ and $\mathbf{u}_l^{(\mathcal{Y})} \in \mathbf{U}^{(\mathcal{Y})}$. These features are used to define costs of an optimization problem, which we minimize subject to three conditions: 1. Neighboring triangles are matched to neighboring triangles. 2. Every triangle of the full shape is matched at most once. 3. Every triangle of the partial shape is matched exactly once. 
    }
    \label{fig:overview_opt}
\end{figure*}

\section{Method}
In Figure~\ref{fig:overview_opt} we show an overview of our method. Our key idea is a novel integer linear programming formalism for geometrically consistent and surjective partial shape matching. For defining a matching cost function we leverage state-of-the-art learning-based features.  
We provide an overview of all needed symbols with short descriptions  in Table~\ref{table:notation}. In the following we introduce the notation to set the stage for a detailed description of our geometrically consistent partial shape matching approach.

\begin{table}[h!]
\small\centering
	\begin{tabularx}{\columnwidth}{lp{5.6cm}}
        \toprule
        \textbf{Symbol} & \textbf{Description} \\
        \toprule
        $\mathcal{X}=\bigl(\mathbf{V}^{(\mathcal{X})},\mathbf{F}^{(\mathcal{X})}\bigr)$&3D shape (triangle mesh)\\
        $|\mathbf{V}^{(\mathcal{X})}|,|\mathbf{E}^{(\mathcal{X})}|,|\mathbf{F}^{(\mathcal{X})}|$&Number of vertices, edges, triangles\\
        $\mathbf{V}^{(\mathcal{X})}\subset\mathbb{R}^3$& Shape vertices \\
        $\mathbf{E}^{(\mathcal{X})}$& (Directed) edges \\
        $\mathbf{\bar{E}}^{(\mathcal{X})}=\mathbf{V}^{(\mathcal{X})}\cup\mathbf{E}^{(\mathcal{X})}$& Extended edges\\
        $\mathbf{F}^{(\mathcal{X})}$& (Oriented) triangular faces \\
        $\mathbf{\bar{F}}^{(\mathcal{X})}=\mathbf{\bar{E}}^{(\mathcal{X})}\cup\mathbf{F}^{(\mathcal{X})}$& Extended triangles\\
        $\mathcal{O}^{(\mathcal{X})}$& Orientation operator\\
        $\bm{\pi}^{(\mathcal{X})}$& Projection operator\\
        $\mathbf{W}^{(\mathcal{X})}\subset\mathbb{R}^{|\mathbf{V}^{(\mathcal{X})}| \times d }$& Vertex-wise features\\
      $\mathbf{U}^{(\mathcal{X})}\subset\mathbb{R}^{|\mathbf{F}^{(\mathcal{X})}| \times d }$& Triangle-wise features\\
        $\mathbf{E}$& Product edges \\
        $\mathbf{F}$& Product triangles\\
        $\mathbf{\Pi}\in\{0,1\}^{|\mathbf{F}|}$& Indicator vector of product space\\
        $\bm{\partial}$& Boundary operator\\
        $\mathbf{C}\in\mathbb{R}^{|\mathbf{F}|}$& Cost vector\\
        \bottomrule
	\end{tabularx}
	\caption{Summary of our notation used in this paper. 
	}
	\label{table:notation}
\end{table}

\paragraph{Notation.}
Consider the oriented, manifold shape $\mathcal{X}=\bigl(\mathbf{V}^{(\mathcal{X})},\mathbf{F}^{(\mathcal{X})}\bigr)$ with vertices $\mathbf{V}^{(\mathcal{X})}:=\bigl\{\mathbf{v}^{(\mathcal{X})}_i\in\mathbb{R}^3|1\leq i \leq |\mathbf{V}^{(\mathcal{X})}|\bigr\}$ and triangles $\mathbf{F}^{(\mathcal{X})}\subset\mathbf{V}^{(\mathcal{X})}\times\mathbf{V}^{(\mathcal{X})}\times\mathbf{V}^{(\mathcal{X})}$.
Let $\mathbf{E}^{(\mathcal{X})}\subset\mathbf{V}^{(\mathcal{X})}\times\mathbf{V}^{(\mathcal{X})}$ be the set of (oriented) edges of the triangles $\mathbf{F}^{(\mathcal{X})}$. A (non-boundary) edge $(\mathbf{v}^{(\mathcal{X})}_i,\mathbf{v}^{(\mathcal{X})}_j)\in \mathbf{E}^{(\mathcal{X})}$ is shared by its two neighboring triangles. Since shape $\mathcal{X}$ is oriented, (non-boundary) edges come in pairs with complementary orientations in  adjacent triangles, denoted by $(\mathbf{v}^{(\mathcal{X})}_i, \mathbf{v}^{(\mathcal{X})}_j)$ and $(\mathbf{v}^{(\mathcal{X})}_j, \mathbf{v}^{(\mathcal{X})}_i)$, respectively.
\begin{figure}
    \centering    
    \includegraphics[width=0.6\linewidth]{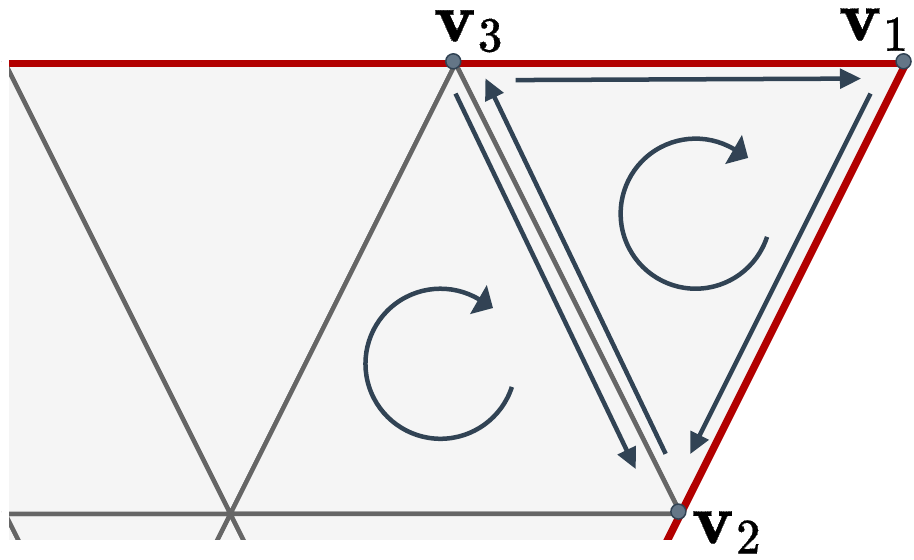}
    \caption{We visualize the \textbf{edge orientation} of an example shape, where boundaries are shown in red. The interior edge $(\mathbf{v_2}, \mathbf{v_3})$ is also present in opposite orientation $(\mathbf{v_3}, \mathbf{v_2})$, while boundary edges only exists in one orientation, e.g.~$(\mathbf{v_1}, \mathbf{v_2})$ and $(\mathbf{v_3},\mathbf{v_1})$.
    }
    \label{fig:orientation}
\end{figure}
For a shape without boundary this means that we encounter every edge in exactly two orientations.
For a shape with boundary we encounter both orientations only for edges which lie in the interior of the shape. 
For a boundary edge we encounter just one orientation as there is no second adjacent triangle to this edge.
See Figure~\ref{fig:orientation} for an illustration.

\begin{figure}
    \centering
\includegraphics[width=1\linewidth]{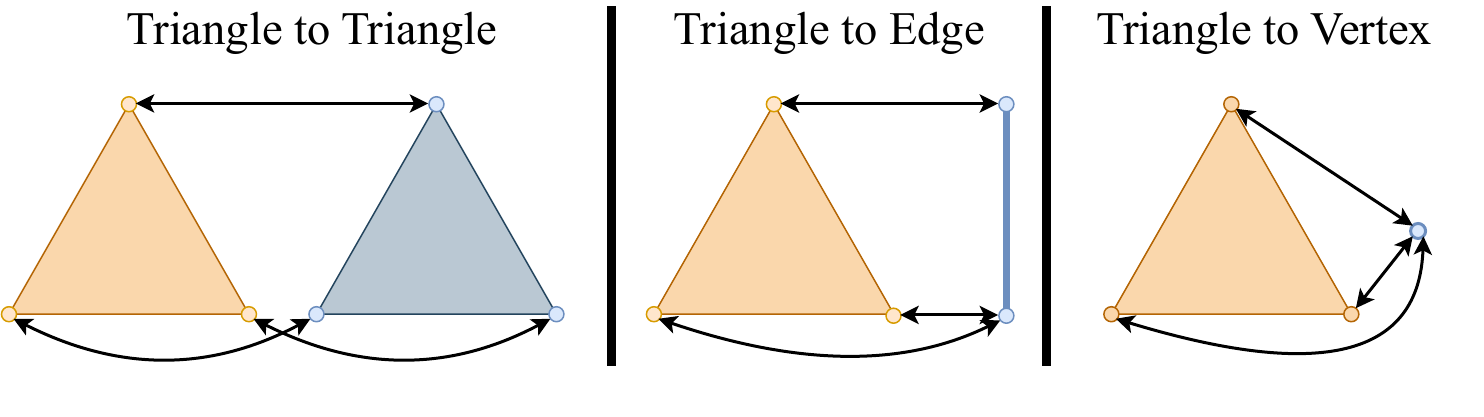}
    \caption{We show different \textbf{possible triangle matchings}. Degenerate triangles allow to model triangle-triangle, triangle-edge and triangle-vertex matchings using a unified formalism.
    }
    \label{fig:possible_triangles}
\end{figure}

We extend the set of (non-degenerate) edges by degenerate edges (i.e.~a vertex can be seen as self-edge) and it as $\mathbf{\bar{E}}^{(\mathcal{X})}=\mathbf{V}^{(\mathcal{X})}\cup\mathbf{E}^{(\mathcal{X})}$. 
Similarly, we extend the set of (non-degenerate) triangles $\mathbf{\bar{F}}^{(\mathcal{X})}=\mathbf{\bar{E}}^{(\mathcal{X})}\cup\mathbf{F}^{(\mathcal{X})}$ with degenerate triangles (edges and vertices).
With this representation, it is possible to treat vertices as edges, and edges and vertices as triangles, so that triangles can not only be matched to (non-degenerate) triangles but also to edges and vertices (cf. Figure~\ref{fig:possible_triangles}).
This is required to account for stretching and compression of triangles (e.g.~to handle non-rigid deformations and discretization differences between both shapes).

We denote by $\mathcal{O}^{(\mathcal{X})}:\mathbf{E}^{(\mathcal{X})}\times \mathbf{F}^{(\mathcal{X})}\to \{-1,0,1\}$ the orientation operator between an edge $\mathbf{e}=(\mathbf{v}_1',\mathbf{v}_2')$ and a triangle $\mathbf{f}=(\mathbf{v}_1,\mathbf{v}_2,\mathbf{v}_3)$, which is defined as: 
    \begin{equation}\label{eq:orientationoperator}
        \mathcal{O}^{(\mathcal{X})}:(\mathbf{e},\mathbf{f})\mapsto\begin{cases}
    \phantom{-}1 & \text{if}~~\mathbf{e}\in\{(\mathbf{v}_1,\mathbf{v}_2),(\mathbf{v}_2,\mathbf{v}_3),(\mathbf{v}_3,\mathbf{v}_1)\} \\
    -1 & \text{if}~~\mathbf{e}\in\{(\mathbf{v}_1,\mathbf{v}_3),(\mathbf{v}_3,\mathbf{v}_2),(\mathbf{v}_2,\mathbf{v}_1)\} \\
    \phantom{-}0 &\text{else}. 
    \end{cases}
    \end{equation}
The orientation operator quantifies whether the edge $\mathbf{e}$ is adjacent to the triangle  $\mathbf{f}$ and encodes their orientation relation.

\subsection{Geometrically Consistent Partial Matching}

We consider a pair of manifold and oriented 3D shapes $\mathcal{X}, \mathcal{Y}$, where w.l.o.g.~we assume that the full shape $\mathcal{X}$ is closed (i.e.~no boundary) and the partial shape $\mathcal{Y}$ is open (i.e.~with boundary/boundaries). Our objective is to find correspondences between (potentially degenerate) triangles of $\mathcal{X}$ and $\mathcal{Y}$.
Our partial matching formulation builds upon and extends the full shape matching formalism of Windheuser \etal~\cite{windheuser2011geometrically, windheuser2011large, Schmidt-et-al-14}. 

We consider product edges $\mathbf{E}:=\bigl(\mathbf{\bar{E}}^{(\mathcal{X})}\times\mathbf{\bar{E}}^{(\mathcal{Y})}\bigr)\setminus\bigl(\mathbf{V}^{(\mathcal{X})}\times\mathbf{V}^{(\mathcal{Y})}\bigr)$ ($\setminus$ denotes set difference), where every non-degenerate edge in $\mathcal{X}$ is combined with every (non-)~degenerate edge in $\mathcal{Y}$ and vice versa. 
Similar we define product triangles $\mathbf{F}:=\bigl(\mathbf{\bar{F}}^{(\mathcal{X})}\times\mathbf{\bar{F}}^{(\mathcal{Y})}\bigr)\setminus\bigl(\mathbf{\bar{E}}^{(\mathcal{X})}\times\mathbf{\bar{E}}^{(\mathcal{Y})}\bigr)$, where every non-degenerate triangle in $\mathcal{X}$ is combined with every (non-)degenerate triangle in $\mathcal{Y}$ and vice versa. 
We represent a correspondence map $\mathbf{\Pi}\in\{0,1\}^{|\mathbf{F}|}$ as an indicator vector, where values of $\mathbf{\Pi}_i=1$ denote a match between the two triangles $\mathbf{f}_k^{(\mathcal{X})}$ and $\mathbf{f}_l^{(\mathcal{Y})}$ that constitute the product triangle $\mathbf{f}_i=(\mathbf{f}_k^{(\mathcal{X})},\mathbf{f}_l^{(\mathcal{Y})})\in\mathbf{F}$. In our integer linear program we aim to find the entries in $\mathbf{\Pi}$ and consequently
triangle-triangle (degenerate or non-degenerate) correspondences between both shapes.
\paragraph{Cost Function.}
To decide which matchings are favorable, we define a cost function based on an off-the-shelf shape feature extractor that returns a feature vector $\mathbf{w}^{(\mathcal{X})}_i$ with  feature dimensionality $d$ for every vertex $\mathbf{v}^{(\mathcal{X})}_i$ of a given shape $\mathcal{X}$,
which can be stacked into the matrix $\mathbf{W}^{(\mathcal{X})}$ containing all feature descriptors of shape $\mathcal{X}$. The feature vector $\mathbf{u}_k^{(\mathcal{X})} \in \mathbf{U}^{(\mathcal{X})}$ 
for triangle $\mathbf{f}_k$ is simply defined as the mean of the features of respective triangle vertices.
The cost $c_i$ of a product triangle $\mathbf{f}_i=(\mathbf{f}_k^{(\mathcal{X})},\mathbf{f}_l^{(\mathcal{Y})})\in\mathbf{F}$ is defined by the L2 norm of the features $\mathbf{u}_k^{(\mathcal{X})}, \mathbf{u}_l^{(\mathcal{Y})}$, i.e.~$c_i = ||\mathbf{u}_k^{(\mathcal{X})} - \mathbf{u}_l^{(\mathcal{Y})}||_2$.

\paragraph{Geometric Consistency Constraints.}
While minimizing our cost function we want to ensure surjectivity, as we expect that the partial shape $\mathcal{Y}$ is fully represented by the full shape $\mathcal{X}$, but the full shape $\mathcal{X}$ is only partly represented by the partial shape $\mathcal{Y}$.
Hence, every (non-degenerate) triangle of the partial shape should be matched exactly once, 
and every (non-degenerate) triangle of the full shape should be matched at most once.
Similar to previous works~\cite{windheuser2011geometrically},
we define this requirement via the projection matrices $\bm{\pi}^{(\mathcal{X})}\in\{0,1\}^{|\mathbf{F}^{(\mathcal{X})}|\times|\mathbf{F}|}$ as
\begin{equation}
        \bm{\pi}^{(\mathcal{X})}_{i,j}:=\begin{cases}
        1 & \text{if}\quad \mathbf{f}_j \text{ contains } \mathbf{f}_i^{(\mathcal{X})} \\
        0 &\text{else.} 
        \end{cases}
\end{equation}
$\bm{\pi}^{(\mathcal{Y})}\in\{0,1\}^{|\mathbf{F}^{(\mathcal{Y})}|\times|\mathbf{F}|}$ is defined analogously, and the surjectivity constraints are defined as
\begin{equation}
\label{eq:op-proj}
    \bm{\pi}^{(\mathcal{X})} \mathbf{\Pi}\leq \mathbf{1},\quad
    \bm{\pi}^{(\mathcal{Y})} \mathbf{\Pi} = \mathbf{1},
\end{equation}
where $\mathbf{1}$ is a vector with all ones.

Further, we want to ensure geometric consistency. Geometric consistency ensures that neighboring elements (triangles, edges, vertices) in shape $\mathcal{X}$ are consistently matched to neighboring elements in $\mathcal{Y}$.
For that purpose, we define $\mathbf{\mathring{E}}\subset\mathbf{E}$ as the set of interior product edges, 
where we exclude all boundary edges in $\mathbf{\bar{E}}^{(\mathcal{Y})}$ of the shape $\mathcal{Y}$.

Moreover, we define the boundary operator $\bm{\partial}\in\{-1,0,1\}^{|\mathbf{\mathring{E}}|\times|\mathbf{F}|}$ using the \emph{product orientation operator} $\mathcal{O}$ as
\begin{equation}
\label{eq:oriention-prod}
\partial_{i,j}:=\mathcal{O}(\mathbf{e}_i,\mathbf{f}_j),
\end{equation}
which assigns an orientation $\{-1,1\}$ to interior product edges $\mathbf{e}_i\in\mathbf{\mathring{E}}$ that coincide with product triangles $\mathbf{f}_j\in\mathbf{F}$. To obtain the product orientation operator $\mathcal{O}$, the (non-product) orientation operator from Eq.~\eqref{eq:orientationoperator} is applied to edges $\mathbf{e}_i$ and triangles $\mathbf{f}_j$ in the $4$-dimensional 
product space (cf.~\cite{windheuser2011geometrically} for further insights). In Eq.~\eqref{eq:oriention-prod} we implicitly extend the orientation operator of a single shape to elements in the product space.
By enforcing $\bm{\partial} \mathbf{\Pi} = \boldsymbol{0}$, we can ensure geometric consistency for all (matched) interior product edges.

By combining the previously introduced components, our final optimization problem reads
    \begin{align}
        ~&~\underset{\mathbf{\Pi} \in \{0, 1\}^{|F|}}{\min} \bigl\langle\mathbf{C},\mathbf{\Pi}\bigr\rangle \\
        ~&~\text{ s.t. } \bm{\partial} \mathbf{\Pi} = \mathbf{0}, \bm{\pi}^{(\mathcal{X})} \mathbf{\Pi} \leq \mathbf{1}, \bm{\pi}^{(\mathcal{Y})}\mathbf{\Pi} = \mathbf{1}. \nonumber
    \end{align}
We use YALMIP~\cite{Lofberg2004} with Gurobi~\cite{gurobi}  as branch and bound solver
to solve our Integer Linear Program to global optimality.

\subsection{Coarse-to-Fine Optimization Scheme}\label{sec:coarse_to_fine}
\begin{figure}
    \centering
    \includegraphics[width=\linewidth]{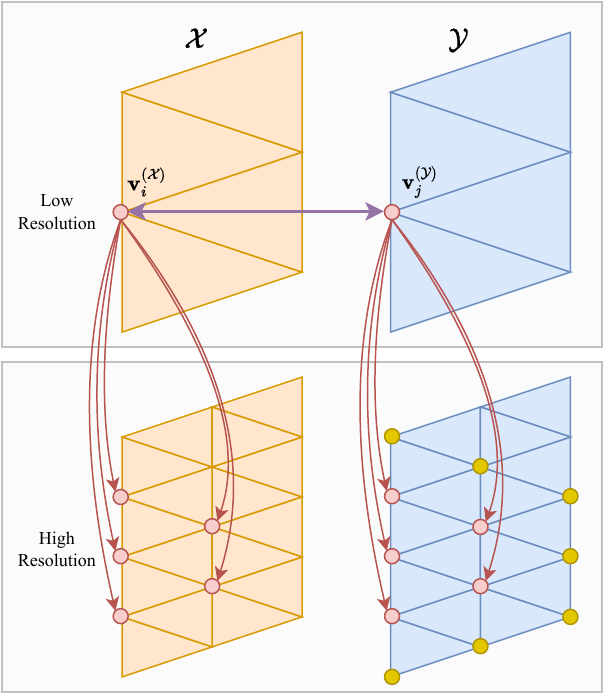}
    \caption{We use a \textbf{coarse-to-fine approach} to solve higher resolution problems. Based on the low resolution matching \mbox{$\mathbf{v}^{(\mathcal{X})}_i \leftrightarrow \mathbf{v}^{(\mathcal{Y})}_j$ (top)}, 
     all corresponding high resolution vertices of $\mathcal{X}$  corresponding to $\mathbf{v}^{(\mathcal{X})}_i$ (red, bottom) can be matched to all corresponding high resolution vertices of $\mathcal{Y}$  corresponding to $\mathbf{v}^{(\mathcal{Y})}_j$ (red) and their 1-ring neighborhood (yellow). 
    }
    \label{fig:method_coarse_to_fine}
\end{figure}
Similar to~\cite{Windheuser-et-al-sgp11} we use a coarse-to-fine scheme to solve the underlying problem for higher resolution shapes. 
We first solve a low resolution shape matching problem to global optimality and use this solution to narrow down the search space for higher resolutions. 
Given a globally optimal matching $\mathbf{v}^{(\mathcal{X})}_i \leftrightarrow \mathbf{v}^{(\mathcal{Y})}_j$ for low resolution shapes, the optimal correspondence of $\mathbf{v}^{(\mathcal{X})}_i$ in the higher resolution shape $\mathcal{Y}$ must lie in the neighborhood of $\mathbf{v}^{(\mathcal{Y})}_j$ (see Figure~\ref{fig:method_coarse_to_fine} for an illustration). 
In practice, we restrict the neighborhood to be the  1-ring around $\mathbf{v}^{(\mathcal{Y})}_j$, which we empirically verified to be efficient and accurate, thanks to the high quality globally optimal correspondences in the low resolution shapes found by our method. Note that the radius of the search neighborhood (\emph{i.e.}~number of rings) is a hyper-parameter that can be adjusted according to the shape discretization.

\section{Experimental Results}
In the following we experimentally evaluate our proposed formalism. To this end, we show that our formalism outperforms existing geometrically consistent algorithms that require to close all holes when addressing partial shape matching. Additionally, we use our method to address limitations of current SOTA shape matching methods related to the lack of geometric consistency in the matchings.

Throughout all experiments we utilize the pre-trained state-of-the-art deep learning-based feature embeddings from~\cite{cao2023unsupervised}.

\subsection{Datasets}
Our experiments use the two \textsc{SHREC16 Partial} datasets \textsc{Cuts} and \textsc{Holes}~\cite{cosmo2016shrec}.
Each of the datasets contains eight animal/human classes in different poses. 
A class contains one full shape in a neutral pose used as a template shape, and multiple partial shapes.
The goal is to find correspondences between each partial shape and the template. 
The \textsc{Cuts}  dataset contains partial shapes with a single clean cut, whereas the holes dataset contains partial shapes with multiple irregular cuts and holes, forming a more challenging setting. 
We evaluate our experiments on the test set of the \textsc{SHREC16 Partial} dataset, which contains 200 shapes in the \textsc{Cuts} dataset and 200 shapes in the \textsc{Holes} dataset.
Given this dataset we generate an edge/vertex manifold~\cite{de2003representation} dataset that we use for our evaluation. 
We refer to the supplementary material for  more information about the dataset generation and processing.

\begin{figure*}[htpb]
\centering
\begin{subfigure}[b]{0.45\linewidth}
\centering
  \includegraphics{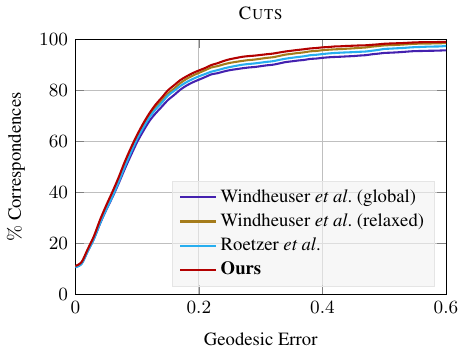}
\end{subfigure}
\hfill
\begin{subfigure}[b]{0.45\linewidth}
\centering
  \includegraphics{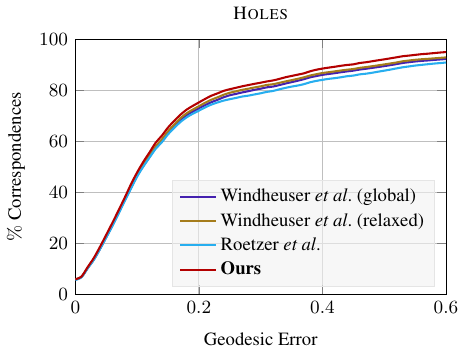}
\end{subfigure}
\caption{
Our method outperforms other methods that \textbf{ensure geometric consistency} on the \textsc{Cuts} and \textsc{Holes} datasets in terms of \textbf{geodesic error}.
To address partial shape matching with competing methods we first
close holes in the shape. We then solve the problem either to global optimality (Windheuser \etal~\cite{windheuser2011geometrically} (global)) or with an LP relaxation (Roetzer \etal~\cite{roetzer2022scalable}, Windheuser \etal~\cite{windheuser2011geometrically} (relaxed)).
}
\label{fig:geo_error_curve}
\end{figure*}

\subsection{Evaluation Metrics}
We evaluate the geodesic error of the correspondences using the Princeton Protocol~\cite{kim2011blended}, where the geodesic error between the ground truth and the computed correspondences is normalized by the shape diameter, i.e.~the square root of the area of the full shape. 
Our method obtains \emph{triangle to triangle} correspondences (potentially involving degenerate triangles) that may involve  one-to-many correspondences between \emph{vertices}. 
To compare with algorithms that return vertex-wise point-to-point maps, for every point we consider the mean geodesic error of all points in correspondence. 
To evaluate whether a matching is smooth, we use the conformal distortion metric~\cite{hormann2000mips} that evaluates the local consistency of triangles after matching, see~\cite[Eq.~(3)]{hormann2000mips} for the definition.

\subsection{Comparison to Geometrically Consistent \\ Algorithms}

In the recent work on geometrically consistent shape matching~\cite{roetzer2022scalable}, the authors show that it possible to generate geometrically consistent matchings between a partial and full shape by closing the holes of the partial shapes  beforehand. 
To compare such methods with our method, we therefore close holes in the partial shape by adding an extra vertex in the middle of every hole and then connect all boundary vertices of this hole with the extra vertex, such that we obtain a closed manifold oriented partial shape. For our experiment we use shapes that are downsampled to 100 faces. 
For fairness, we also provide the competing methods with our learning-based feature energy (otherwise the other methods perform way worse, see supp.). We set a maximum time budget of two hours to find correspondences for each shape pair. As the extra vertices are not present in the original shape, for matching holes we assign the mean feature descriptor over all boundary vertices of the corresponding hole. 
The extra vertices are not used for evaluation as they do not have any ground truth correspondence.

We compare our algorithm (which does not require hole closing) to  existing geometrically consistent algorithms~\cite{windheuser2011geometrically, roetzer2022scalable}, which however require watertight shapes as input (hence our pre-processing step to close holes as described above).
We consider different variants of these methods: solving to global optimality via branch and bound, and 
linear programming relaxations.

We show the geodesic errors in Figure~\ref{fig:geo_error_curve}. Our method outperforms the other methods in both datasets (\textsc{Cuts} and Holes). 

Additionally  in Table~\ref{tab:consistent_sol_comp} we show that our method is able to successfully solve all matching instances, whereas all competitors have failures, i.e.~no geometrically consistent matching is found within the time budeget.
The additional step of closing holes adds complexity to the problem, which empirically leads to worse performance, especially in the \textsc{Holes} dataset.

\begin{table}
  \centering
  \begin{tabular}{@{}lcl@{}}
      \toprule
 Method & \textsc{Cuts}  & \textsc{Holes}  \\
     \midrule
Windheuser \etal~\cite{windheuser2011geometrically} (global) & 6 & 9 \\ 
Windheuser \etal~\cite{windheuser2011geometrically} (relaxed) & 1 & 9 \\  
Roetzer \etal~\cite{roetzer2022scalable} & 8 & 20 \\
Ours & \textbf{0} & \textbf{0} \\
      \bottomrule
  \end{tabular}
  \caption{ 
  We compare \textbf{the number of failed }
  \textbf{matchings} of downsampled shapes in coarse resolution (100 faces) on the \textsc{SHREC16 Partial} datasets \textsc{Cuts} and \textsc{Holes}. 
  A failed matching means that no geometrically consistent matching is found within the maximum time budget.
  In comparison to the other methods, our method does not fail to compute a geometrically consistent matching between a partial and full shape. 
  }
  \label{tab:consistent_sol_comp}
\end{table}
\newcommand{\includeCoarseToFine}[3]{
        \includegraphics[width= 0.19\linewidth]{pictures/coarse_to_fine/#2_#3_#1.png}
}
\newcommand{\includeCoarseToFineRow}[2]{
\includeCoarseToFine{200}{#1}{#2} &\includeCoarseToFine{400}{#1}{#2} &\includeCoarseToFine{600}{#1}{#2} &\includeCoarseToFine{800}{#1}{#2}&\includeCoarseToFine{1000}{#1}{#2}
}

\begin{figure*}[htpb]
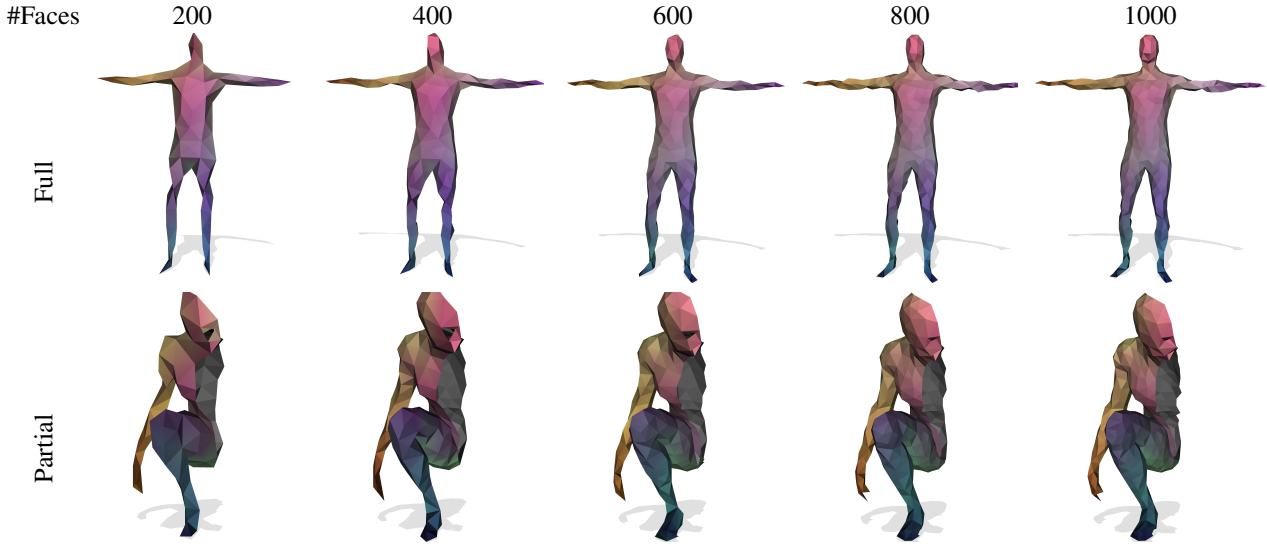

    \centering
    \setlength\tabcolsep{-3pt} 
        \begin{tabular}{c c c c c c}
            \centering
             \#Faces &200 &  400 & 600 & 800 & 1000\\
             \multirow{2}{*}[4em]{\rotatebox{90}{Full}} & \includeCoarseToFineRow{michael}{michael_cuts_michael_shape_6} \\
             \multirow{2}{*}[4em]{\rotatebox{90}{Partial}} &\includeCoarseToFineRow{cuts_michael_shape_6}{michael_cuts_michael_shape_6}
        \end{tabular}
        \vspace{-3mm}
    \caption{Our \textbf{coarse-to-fine scheme}. Initially, we perform a coarse matching (full shape with 200 faces, partial shapes are decimated accordingly).
    Based on this coarse matching we prune the search space for subsequent higher resolution matchings. 
    }
    \label{fig:coarseToFine}
\end{figure*}

\subsection{Coarse-to-fine Matching}
To be able to solve higher resolution shapes we use the coarse-to-fine scheme presented in Section~\ref{sec:coarse_to_fine}.
For the lowest resolution we start with a resolution of 200 faces (if this is solvable in reasonable time of two hours). 
Otherwise we start with 100 faces as lowest resolution.
In our experiments only a single matching instance required to start at the lower resolution of 100 faces.
We solve every subsequent step to global optimality in the reduced search space.
We increase the size of the finest problem to 1000 faces (in the full shape).
We show one example of the coarse-to-fine scheme in Figure~\ref{fig:coarseToFine}.

\subsection{Comparison to SOTA Learning-Based Shape Matching}
To demonstrate that our proposed method can  improve upon  learning-based shape matching methods, we compare it against the SOTA method~\cite{cao2023unsupervised}. Since the method proposed in~\cite{cao2023unsupervised} substantially outperforms all other (unsupervised) methods in their experiments, in our evaluation  we restrict ourselves exclusively to a comparison with~\cite{cao2023unsupervised}.

In our experiments, we additionally consider the generalization setting of~\cite{cao2023unsupervised}, where the feature extractor is pretrained on the other dataset: the \textsc{Cuts} dataset pretrained on the \textsc{Holes}, and the \textsc{Holes} dataset pretrained on \textsc{Cuts}.
In Table~\ref{tab:geoerror} we show that our algorithm can maintain (\textsc{Cuts} trained on \textsc{Cuts}) or even improve (all other datasets) the geodesic error on the \textsc{SHREC16 Partial} dataset, while guaranteeing geometric consistency.

\begin{table}
  \centering
  \begin{tabular}{@{}lcl@{}}
      \toprule
 Dataset & \cite{cao2023unsupervised} & Ours \\
     \midrule
\textsc{Cuts} & \textbf{4.41} & 4.45 \\  
\textsc{Holes} & 9.51 & \textbf{8.19} \\
\textsc{Cuts} trained on \textsc{Holes} & 6.10 & \textbf{5.58} \\
\textsc{Holes} trained on \textsc{Cuts} & 13.87 & \textbf{11.40} \\
      \bottomrule
  \end{tabular}
  \caption{\textbf{The geodesic error ($\times$100)} of both methods is low for the \textsc{Cuts} dataset on downsampled shapes of 1000 faces.
  Our method outperforms the current unsupervised SOTA shape matching method in the three other settings. Moreover, our method guarantees geometric consistency. }
  \label{tab:geoerror}
\end{table}

\begin{figure}[htbp]
    \begin{tabular}{ll}

      \includegraphics{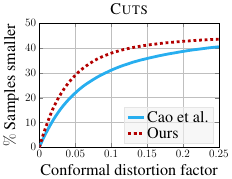}   & \includegraphics{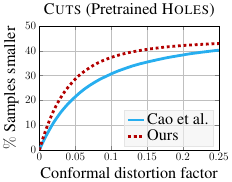} \\
      \includegraphics{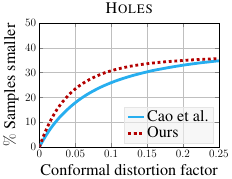}    &  \includegraphics{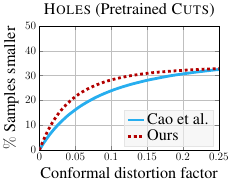}
    \end{tabular}
    
    \caption{We analyze the \textbf{cumulative conformal distortion} of our method in comparison to the state-of-the-art learning-based method~\cite{cao2023unsupervised}. Our method is much smoother in all four settings.
    }
    \label{fig:smoothness}
\end{figure}

\newcommand{\includeMorphShape}[2]{
        \includegraphics[width= 0.19\linewidth]{pictures/morphing/cuts_michael_shape_8_michael_#2#1.png}
}

\newcommand{\includeMorphShapeDog}[2]{
        \includegraphics[width= 0.18\linewidth]{pictures/morphing/cuts_dog_shape_10_dog_#2#1.png}
}

\newcommand{\includeMorphRow}[1]{
\includeMorphShape{01}{#1} &\includeMorphShape{03}{#1} &\includeMorphShape{05}{#1} &\includeMorphShape{07}{#1} &\includeMorphShape{10}{#1}
}

\newcommand{\includeMorphRowDog}[1]{
\includeMorphShapeDog{01}{#1} &\includeMorphShapeDog{03}{#1} &\includeMorphShapeDog{05}{#1} &\includeMorphShapeDog{07}{#1} &\includeMorphShapeDog{10}{#1}
}

\begin{figure*}[htbp]
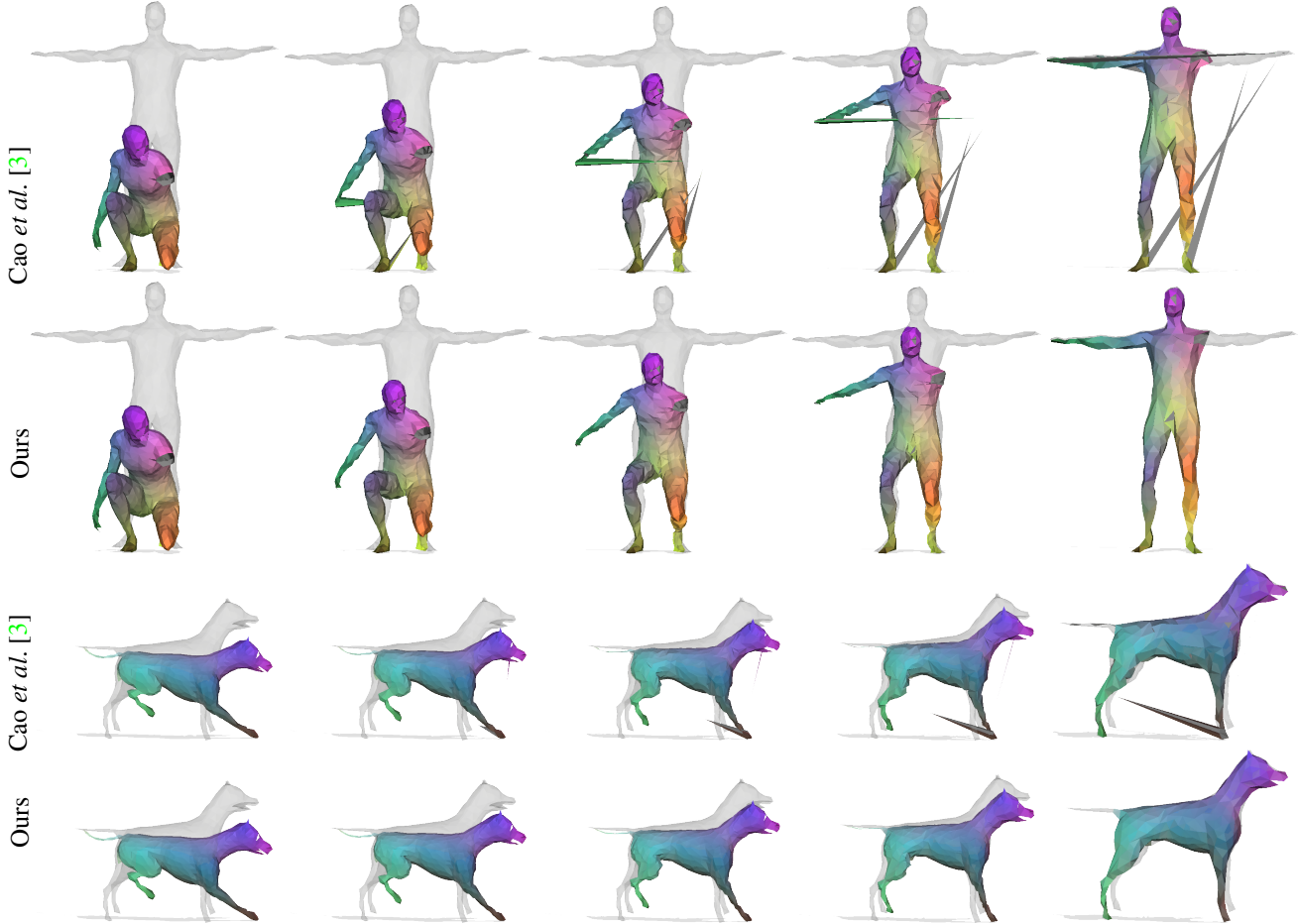

    \centering
    \setlength\tabcolsep{0pt}
        \begin{tabular}{cccccc}
            \multirow{2}{*}[4em]{\rotatebox{90}{Cao \etal~\cite{cao2023unsupervised}}} & \includeMorphRow{siggraph} \\
            \multirow{2}{*}[4em]{\rotatebox{90}{Ours}} & 
            \includeMorphRow{normal}\\
            \multirow{2}{*}[4em]{\rotatebox{90}{Cao \etal~\cite{cao2023unsupervised}}} & \includeMorphRowDog{siggraph} \\
            \multirow{2}{*}[4em]{\rotatebox{90}{Ours}} &\includeMorphRowDog{normal}
        \end{tabular}
            \caption{We \textbf{linearly interpolate} \textbf{a partial (color) to a full shape (gray)} (human in the top, animal in the bottom). The interpolation steps are shown from left to right. The SOTA learning-based method~\cite{cao2023unsupervised} results in wrong matches (e.g.~at arms/legs in the top), so that linear shape interpolation results in obvious errors. Our method guarantees geometric consistency and does not suffer from such issues.
            }
            \label{fig:interpolate}
\end{figure*}

A small geodesic error does not imply a smooth matching. As our method ensures geometric consistency we show that it improves the matching smoothness, which is necessary for tasks like texture transfer or interpolation.
To measure the smoothness of a matching we use the conformal distortion~\cite{hormann2000mips}. In Figure~\ref{fig:smoothness} we show the cumulative conformal distortion of our algorithm compared to~\cite{cao2023unsupervised}. Our algorithm returns smoother results in all four settings compared to the baseline~\cite{cao2023unsupervised}. In Figure~\ref{fig:interpolate} we show two qualitative examples, where our method outperforms the baseline  in a linear interpolation task. Due to geometric consistency, our algorithm matches neighboring triangles to neighboring triangles, while~\cite{cao2023unsupervised} has a few mismatches. While such mismatches do not substantially affect the geodesic error, they result in obvious interpolation errors. For animated shape morphing examples please refer to the supplementary material.

\section{Discussion and Limitations}
Our method shares limitations with existing geometrically consistent shape matching algorithms~\cite{windheuser2011large,roetzer2022scalable}, which are limited in terms of scalability due to the extremely large integer linear program. Yet, while competing methods often lead to failed matchings (see \cref{tab:consistent_sol_comp}), our method is able to successfully find feasible and globally optimal solutions at the coarsest scale, which in turn makes it possible to gradually expand coarse solutions  to finer scales -- failures at a coarse shape resolution, as encountered with the other methods, will directly break any coarse-to-fine scheme.

Recent learning based shape matching methods~\cite{attaiki2021dpfm,cao2023unsupervised}  can handle shapes with thousands of vertices, however, at the expense of lacking geometric consistency.  In contrast, our method handles shapes with hundreds of vertices, but guarantees geometric consistency. We consider our work as a first step towards enabling the best of both worlds.

Another aspect that opens up interesting avenues for future work is to address the matching of pairs of partial shapes when overlapping regions are not known a-priori.

\section{Conclusion}
We have presented a novel formalism for partial to full shape matching that enforces geometric consistency based on linear constraints.
To optimize the resulting large and difficult integer linear program we proposed to combine SOTA deep shape features with a coarse-to-fine scheme that gradually expands a globally optimal low resolution map to higher resolutions.
Compared to previous geometrically consistent shape matching methods, our method can directly address partial shape matching without the need of filling holes with a dummy geometry.
We experimentally demonstrated that our method compares favorably to such methods, and that our method can even substantially improve the smoothness of purely learning-based shape matching approaches. We see our work as an important step towards combining the advantages of data-driven and axiomatic  approaches in order to address more challenging and practically relevant shape matching settings.

{
    \small
    \bibliographystyle{ieeenat_fullname}
    \bibliography{main}
}

\clearpage
\setcounter{page}{1}

\maketitlesupplementary

\begin{figure*}[htpb]
\centering
\begin{subfigure}[b]{0.95\columnwidth}
\centering
  \includegraphics{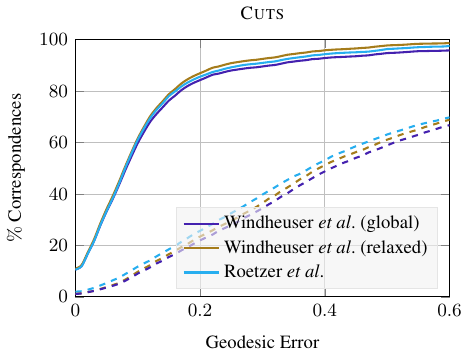}
\end{subfigure}
\hfill
\begin{subfigure}[b]{0.95\columnwidth}
\centering
  \includegraphics{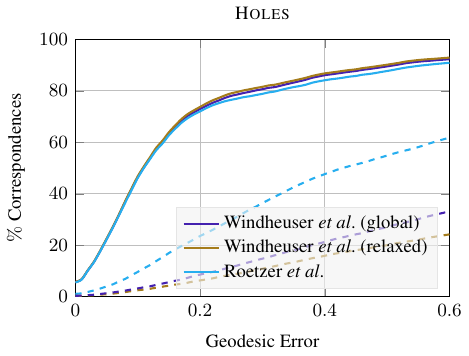}
\end{subfigure}
\caption{
In terms of the \textbf{geodesic error} the comparison methods perform better when utilizing the \textbf{learning-based energy~\cite{cao2023unsupervised} (solid lines)} in contrast to the \textbf{originally used energies~\cite{roetzer2022scalable, windheuser2011geometrically} (dashed lines)} on the \textsc{Cuts} and \textsc{Holes} dataset with downsampled shapes on 100 faces. 
}
\label{fig:geo_error_curve_supp}
\end{figure*}

\section{Data Preparation}
We prepare our data by orienting all normals
and ensuring vertex/edge manifoldness~\cite{de2003representation}. 
Especially in the irregular \textsc{Holes}  
dataset we observe single triangles which are only connected by one vertex to the rest of the shape. 
We remove these triangles automatically by duplicating respective vertices 
and removing unconnected shape parts, such that we end up with a vertex/edge manifold, oriented shape, while we make sure that limited information gets lost (we remove less than $0.6\%$ of all triangles).
For the full shape we additionally close small holes (like the eye sockets),
such that we get a closed, oriented, manifold shape. We end up with an edge and vertex manifold, oriented dataset.

\section{Shape Decimation}
The problem size of our Integer Linear Program 
grows quadratically with the number of faces of the shapes. 
We downsample the input meshes from their original resolution to lower resolution in order to keep the Integer Linear Programs small enough such that we are able to solve them. 
To this end, we use a simple 
edge collapse method,
which merges two vertices into one in every reduction step. 
We iteratively use the decimation steps to keep track of the merged vertices in every single step. 
We use these decimation steps until a specific number of vertices is reached for the full shape $\mathcal{X}$. 
We reduce the partial shape $\mathcal{Y}$ to $2 \cdot A(\mathcal{Y})/A(\mathcal{X}) \cdot |\mathbf{V}^{(\mathcal{X})}|$ number of faces. Here, $A(\cdot)$ is the function that returns the surface area of the respective shape.
By multiplying with $2$ we ensure that we preserve more details of the partial shape on the coarsest stage which is important to make later coarse-to-fine matching computations more robust.

In addition, we transfer the features computed on the high-resolution shape to the low-resolution shape.
To this end, we map each low-resolution vertex to a high-resolution vertex via the identity of the edge-collapse algorithm.
Furthermore, we compute Voronoi areas for each vertex on the low-resolution shape and average features of all vertices on high-resolution shape which lie inside this Voronoi area around the mapped high-resolution vertex.

\section{Morphing Results}
For every animal and human class we provide a linear interpolated morphing example animation in .mp4 format in the supplementary material.

\section{Ablation Studies: Energy Term}
As outlined in the main paper we ensure a fair comparison between our method and the geometrically consistent comparison methods~\cite{windheuser2011geometrically, roetzer2022scalable} by providing the same energy term with learning-based features~\cite{cao2023unsupervised} to all methods. 
We show that this energy term performs better for each individual comparison method than their original energy formulations~\cite{windheuser2011geometrically, roetzer2022scalable} in terms of failed matchings (Table~\ref{tab:failed_matchings_supp}) and geodesic error (Figure~\ref{fig:geo_error_curve_supp}).
We conduct this experiment on both datasets \textsc{Cuts} and \textsc{Holes} on downsampled shapes with 100 faces. 
Roetzer \etal~\cite{roetzer2022scalable} use an improved version of the energy term provided in~\cite{windheuser2011geometrically}  yielding better performance in terms of the geodesic error than the other methods tested with~\cite{windheuser2011geometrically}. 
Still, the new energy term with learning-based features~\cite{cao2023unsupervised} performs substantially better for all methods compared to their original energy formulations~\cite{windheuser2011geometrically, roetzer2022scalable}.

\begin{table}
   \centering
   \begin{tabular}{@{}lcc@{}}
       \toprule
         Method & Original
         & Learned
         \\
     \midrule
      \midrule
         \textsc{Cuts}&& \\
      \midrule
      Windheuser \etal~\cite{windheuser2011geometrically} (global) & 14 & \textbf{6} \\
      Windheuser \etal~\cite{windheuser2011geometrically} (relaxed) & 11 & \textbf{1} \\
      Roetzer \etal~\cite{roetzer2022scalable} & 18& \textbf{8} \\
     \midrule
      \midrule
     \textsc{Holes}&& \\
      \midrule
      Windheuser \etal~\cite{windheuser2011geometrically} (global) & 58 & \textbf{9} \\
      Windheuser \etal~\cite{windheuser2011geometrically} (relaxed) & 79 & \textbf{9} \\
      Roetzer \etal~\cite{roetzer2022scalable} & 38 & \textbf{20} \\
       \bottomrule
   \end{tabular}
   \caption{We show the \textbf{failed matchings} of the comparison methods with the learning-based energy~\cite{cao2023unsupervised} in comparison to their originally used energies~\cite{roetzer2022scalable, windheuser2011geometrically} on the \textsc{Cuts} and \textsc{Holes} dataset with downsampled shapes on 100 faces.}
   \label{tab:failed_matchings_supp}
 \end{table}

\end{document}